\documentclass[conference]{IEEEtran}
\IEEEoverridecommandlockouts

\usepackage{cite}
\usepackage{amsmath,amssymb,amsfonts}
\usepackage{algorithmic}
\usepackage{graphicx}
\usepackage{textcomp}
\usepackage{xcolor}
\usepackage{bm}

\usepackage{caption}
\usepackage[subrefformat=parens]{subcaption}
\def\BibTeX{{\rm B\kern-.05em{\sc i\kern-.025em b}\kern-.08em
    T\kern-.1667em\lower.7ex\hbox{E}\kern-.125emX}}
\begin{document}

\title{Motion ReTouch: Motion Modification \\ Using Four-Channel Bilateral Control

\thanks{This work was supported by JSPS KAKENHI Grant Number 24K00905, JST, PRESTO Grant Number JPMJPR24T3 Japan and JST ALCA-Next Japan, Grant Number JPMJAN24F1. This study was also based on the results obtained from the JPNP20004 project subsidized by the New Energy and Industrial Technology Development Organization (NEDO).}
}

\author{
\IEEEauthorblockN{Koki Inami}
\IEEEauthorblockA{\textit{Intelligent and Mechanical Interaction Systems} \\
\textit{University of Tsukuba}\\
Tsukuba, Japan \\
inami.koki.qy@alumni.tsukuba.ac.jp}
\and
\IEEEauthorblockN{Sho Sakaino}
\IEEEauthorblockA{\textit{Systems and Information Engineering} \\
\textit{University of Tsukuba}\\
Tsukuba, Japan \\
sakaino@iit.tsukuba.ac.jp}
\and
\IEEEauthorblockN{Toshiaki Tsuji}
\IEEEauthorblockA{\textit{Science and Engineering} \\
\textit{Saitama University}\\
Saitama, Japan \\
tsuji@ees.saitama-u.ac.jp}
}

\maketitle

\begin{abstract}
Recent research has demonstrated the usefulness of imitation learning in autonomous robot operation. In particular, teaching using four-channel bilateral control, which can obtain position and force information, has been proven effective. However, control performance that can easily execute high-speed, complex tasks in one go has not yet been achieved. We propose a method called Motion ReTouch, which retroactively modifies motion data obtained using four-channel bilateral control. The proposed method enables modification of not only position but also force information. This was achieved by the combination of multilateral control and motion-copying system. The proposed method was verified in experiments with a real robot, and the success rate of the test tube transfer task was improved, demonstrating the possibility of modification force information.

\end{abstract}


\section{INTRODUCTION}
In recent years, imitation learning~\cite{Hussein_ImitationLearningSurvey}~\cite{fang_SurveyImitationLearning}\cite{Attia_OverviewImitationLearinig}, a learning-based approach that enables robots to imitate human behavior, has been attracting attention. Collecting human motions is called teaching. Some teaching methods have been proposed. The most simple method is kinesthetic teaching~\cite{gavspar_DynamicMovementPrimitives}. Kinesthetic teaching is a method in which a person holds a robot, directly controls it, and has the robot carry out a task to convey the trajectory of the robot's movements. On the other hand, to obtain command values in addition to the robot's trajectory, imitation learning with teleoperation has been proposed~\cite{zhao_ALOHA}\cite{Fu_MobileALOHA}. Imitation learning using four-channel bilateral control has also been proposed as a method for teaching not only position information but also force infomation~\cite{Adachi_imitate_bilate}\cite{Buamanee_BiACT}.

Four-channel bilateral control~\cite{Sakaino_bilate_obliquecordinate} is a method of synchronizing position and force. This method makes it easy to collect human tactile motion data and human operation policy. Furthermore, imitation learning with four-channel bilateral control can cooperate with humans and handle soft and rigid objects with autonomous operation~\cite{sasagawa_cooperation}\cite{Yamane_grasp_hand}\cite{Akagawa_BiIL}.

Imitation learning needs robot motion data without mistakes. This is because incorrect motion data will result in failure motion being learned. However, it is difficult to get a fast or complicated motion using four-channel bilateral control. Although four-channel bilateral control-based imitation learning succeeded in contact-rich tasks, the complexity of the tasks is limited.

In order to solve this problem, post-editing of failed data is required. This is because it is difficult to achieve fast and complex movements in one attempt, but if we can correct failures as necessary, we can generate correct motion data.

Incidentally, multilateral control~\cite{Katsura_Multilateral} has been proposed as a technology to synchronize the positions and forces of three or more robots. In addition, a motion-copying system~\cite{Yokokura_MotionCopyingStability}\cite{igarashi_motioncopy}\cite{Fujisaki_MotionCopying} has been proposed as a technology to reproduce the motion with four-channel bilateral control.

In this paper, we propose a post-editing technology using a four-channel bilateral control called Motion ReTouch which combines multilateral control and a motion-copying system. Specifically, by incorporating pre-correction data into the bilateral control and using multilateral control of three units, the system reflects both past and current operations.

We conducted experiment in the real world using a seven-degree-of-freedom (DoF) robot and compared the success rate of the motion-copying at 3x speed and the motion-copying at 3x speed with Motion ReTouch. We also compared the position and reaction force responses of the two methods. As a result, the success rate of the test tube transfer task was improved by Motion ReTouch. It was also confirmed that the force information had changed.

\begin{figure*}[t]
\centering
\begin{minipage}[b]{0.24\linewidth}
    \centering
    \includegraphics{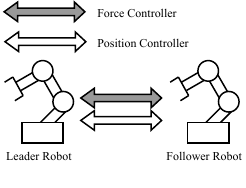}
    \subcaption{4-CH Bilateral Control}
    \label{Fig:illustration_bilate}
\end{minipage}
\begin{minipage}[b]{0.24\linewidth}
    \centering
    \includegraphics{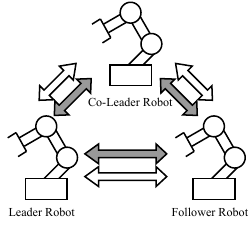}
    \subcaption{Multilateral Control}
    \label{Fig:illustration_multilate}
\end{minipage}
\begin{minipage}[b]{0.24\linewidth}
    \centering
    \includegraphics{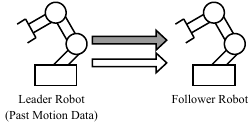}
    \subcaption{Motion Copying System}
    \label{Fig:illustration_motioncopying}
\end{minipage}
\begin{minipage}[b]{0.24\linewidth}
    \centering
    \includegraphics{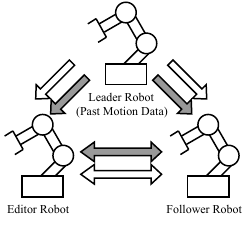}
    \subcaption{Motion ReTouch}
    \label{Fig:illustration_retouch}
\end{minipage}
\caption{Illustration of Controller}
\label{Fig:illustration_all}
\end{figure*}

\section{RELATED WORK}

A large number of methods to teach human skills have been proposed. The most simple method is kinesthetic teaching. It collects the trajectory of the robot as it performs a task. To acquire command values for robots, in addition to the response, methods acquiring motion data by teleoperation have been proposed~\cite{zhao_ALOHA}\cite{Fu_MobileALOHA}. In addition, teleoperation enables the collection of visual information. This is because it avoids an occlusion of the operator and the environment.
A teaching method using virtual reality has also been proposed to make operations more intuitive~\cite{Zhang_ImitationVirtualReality}. These methods allow for the collection of position and visual information but do not include force information. Force information can also be obtained by collecting motion data using four-channel bilateral control~\cite{sasagawa_cooperation}\cite{Yamane_grasp_hand}\cite{Akagawa_BiIL}.

In all imitation learning processes, including four-channel bilateral control-based imitation learning, it is important to obtain correct motion data without errors. However, it is difficult to obtain correct motion data for fast and complex tasks.

Shared autonomy control~\cite{nishimura_HapticSharedControl}\cite{dragan_PolicyBlending}\cite{muelling_SharedAutonomy} is an approach to obtain better motion. Shared autonomy control allows a task to be accomplished with the assistance of a controller. In 4ch bilateral control, it becomes necessary to correct not only position information but also force information. Methods for incorporating shared control into bilateral control have been proposed~\cite{sun_BilateSharedControl}. However, these methods need explicit programming, which spoils the advantage of model-free imitation learning

In addition, interactive imitation learning~\cite{ross_dagger}\cite{kelly_hg-dagger} has been proposed. It is a method of intervening in the behavior of imitation learning. A method for interactive imitation learning using multilateral control has also been proposed~\cite{takahashi_CHG-DAgger}. In this method, force control and acceleration control are removed. Although it is mentioned that training data for the initial model could be obtained through bilateral control, this is not done in practice. In the first place, interactive imitation learning cannot be used without data to train the initial model.

We believe that changing force information is also important, thus in this paper, we propose a method to modify motion data using four-channel bilateral control. This approach allowed for the acquisition of fast and complex motion data, which is difficult to create an initial model.

Incorporating a virtual leader into multilateral control for distance training has also been proposed~\cite{MultiTeletrain_Katsura2008}. We propose to apply this system as a means of data editing.

\section{METHOD}

In this study, we propose Motion ReTouch, a post-editing method using four-channel bilateral control.

Four-channel bilateral control has a position controller and a force controller. The position controller is called the differential mode, and the force controller is called the common mode. They are used to synchronize the positions and forces of the two robots. In the differential mode, the position gap between two robots approaches zero. In the common mode, the force sum of two robots approaches zero. By making the force sum zero, two robots satisfy the law of action and reaction. The illustration of four-channel bilateral control is shown in Fig.~\ref{Fig:illustration_all}\subref{Fig:illustration_bilate}.

Multilateral control~\cite{Katsura_Multilateral} synchronizes the positions and forces of multiple robots. Multilateral control has various combinations of differential mode and common mode. A typical multilateral control system uses the differential mode for every pair of units and the common mode for every unit. The illustration of multilateral control is shown in Fig.~\ref{Fig:illustration_all}\subref{Fig:illustration_multilate}.

Motion-copying system~\cite{Yokokura_MotionCopyingStability}\cite{igarashi_motioncopy}\cite{Fujisaki_MotionCopying} is a method which reproduces the motion data of four-channel bilateral control. When executing four-channel bilateral control, the position and force command values of the follower robot (i.e. the response values of the leader robot) are recorded, and the movement is reproduced by moving the follower robot according to the recorded command values. The illustration of a motion-copying system is shown in Fig.~\ref{Fig:illustration_all}\subref{Fig:illustration_motioncopying}.

We used multilateral control and a motion-copying system to achieve post-editing technology using four-channel bilateral control. We prepared the position and force response values of the leader robot, which needed to be corrected and collected by four-channel bilateral control. We used a follower robot which did tasks. In addition, we incorporated another robot to reflect the motion changes. We called it ``editor robot.''
The illustration of Motion ReTouch is shown in Fig.~\ref{Fig:illustration_all}\subref{Fig:illustration_retouch}.

Differential mode and common mode in Motion ReTouch were defined as follows;
\begin{equation}
    \label{Eq:Differential_Mode_of_Motion_ReTouch}
    \bm{\theta}_{l}^{res} = \bm{\theta}_{f}^{res} = \bm{\theta}_{e}^{res},
\end{equation}
\begin{equation}
    \label{Eq:Common_Mode_of_Motion_ReTouch}
    \bm{\tau}_{l}^{res} + \bm{\tau}_{f}^{res} + \bm{\tau}_{e}^{res} = \bm{0}
\end{equation}
where $\bm{\theta}^{res}$ and $\bm{\tau}^{res}$ represent angle and reaction force response vectors, respectively. The subscripts $\bigcirc_{l}$, $\bigcirc_{f}$, and $\bigcirc_{e}$ represent the leader robot, the follower robot, and the editor robot, respectively.

Because the goal of the differential mode was to make the position deviation zero, it could also be expressed as follows: 

\begin{equation}
\begin{split}
    \label{Eq:Differential_Mode_of_Motion_ReTouch_2}
    \bm{\theta}_{l}^{res} - \bm{\theta}_{f}^{res} = \bm{0},\\
    \bm{\theta}_{f}^{res} - \bm{\theta}_{e}^{res} = \bm{0}, \\
    \bm{\theta}^{res} - \bm{\theta}_{l}^{res} = \bm{0}.
\end{split}
\end{equation}

\begin{figure}[t]
    \centering
    \includegraphics{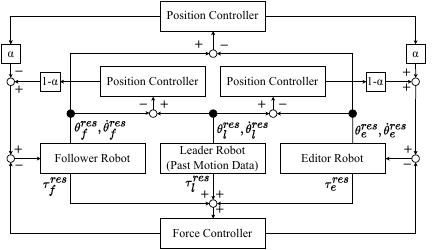}
        \caption{Block Diagram of Motion ReTouch}
        \label{Fig:Block Diagram of Motion ReTouch}
\end{figure}

The block diagram of Motion ReTouch is shown in Fig.~\ref{Fig:Block Diagram of Motion ReTouch}. $\dot{\theta}_{res}$ represents the angler velocity vector. The follower robot
and the editor robot were robots that actually moved, thus they had inputs and outputs. On the other hand, the leader robot only had outputs, because its values were obtained from a record of past motion. All robots were assumed to be acceleration-controlled. The leader robot in four-channel bilateral control and the editor robot in Motion ReTouch should not touch anything other than the hand that was trying to operate it.

First, we considered when no change in motion was necessary. When the editor robot was not subjected to external forces, $F_{e}^{res} = 0$ led to $F_{l}^{res} + F_{f}^{res} = 0$. The follower robot behaved in a way that satisfied the law of action and reaction with the leader robot. Furthermore, the position of the editor robot operated based on the internal division point between the positions of the leader robot and the follower robot. When the positions of the leader robot and the follower robot were synchronized, the positions of the three robots were synchronized. Ideally, the behavior of the follower robot was not affected by the editor robot and was expected to be similar to the motion-copying.

Second, we considered when motion changes were made. When human changes were made to the editor robot, the force of the follower robot approaches $-F_{l}^{res}-F_{e}^{res}$, according to (\ref{Eq:Common_Mode_of_Motion_ReTouch}). Furthermore, the position of the follower robot approaches both the leader and editor robots according to (\ref{Eq:Differential_Mode_of_Motion_ReTouch_2}).
As a result, the position of the follower robot approaches the internal division point between the positions of the leader and editor robots. The ratio of internal division is determined by the parameter $\alpha$ in Fig.~\ref{Fig:Block Diagram of Motion ReTouch}. Basically, it takes values between zero and one. The closer it is to zero, the higher the rate at which past behavior is reproduced. The closer it is to one, the more susceptible it is to changes.

The difference between a typical multilateral control and Motion ReTouch is that the leader robot is the past motion data. In a typical multilateral control, the positions of all robots are always synchronized. On the other hand, the position of the leader robot is unaffected in Motion ReTouch while the positions of the follower and editor robots change.

\section{EXPERIMENTAL DESIGN}
\subsection{ROBOT SYSTEM}

\begin{figure}[t]
    \centering
        \includegraphics[width=0.25\linewidth]{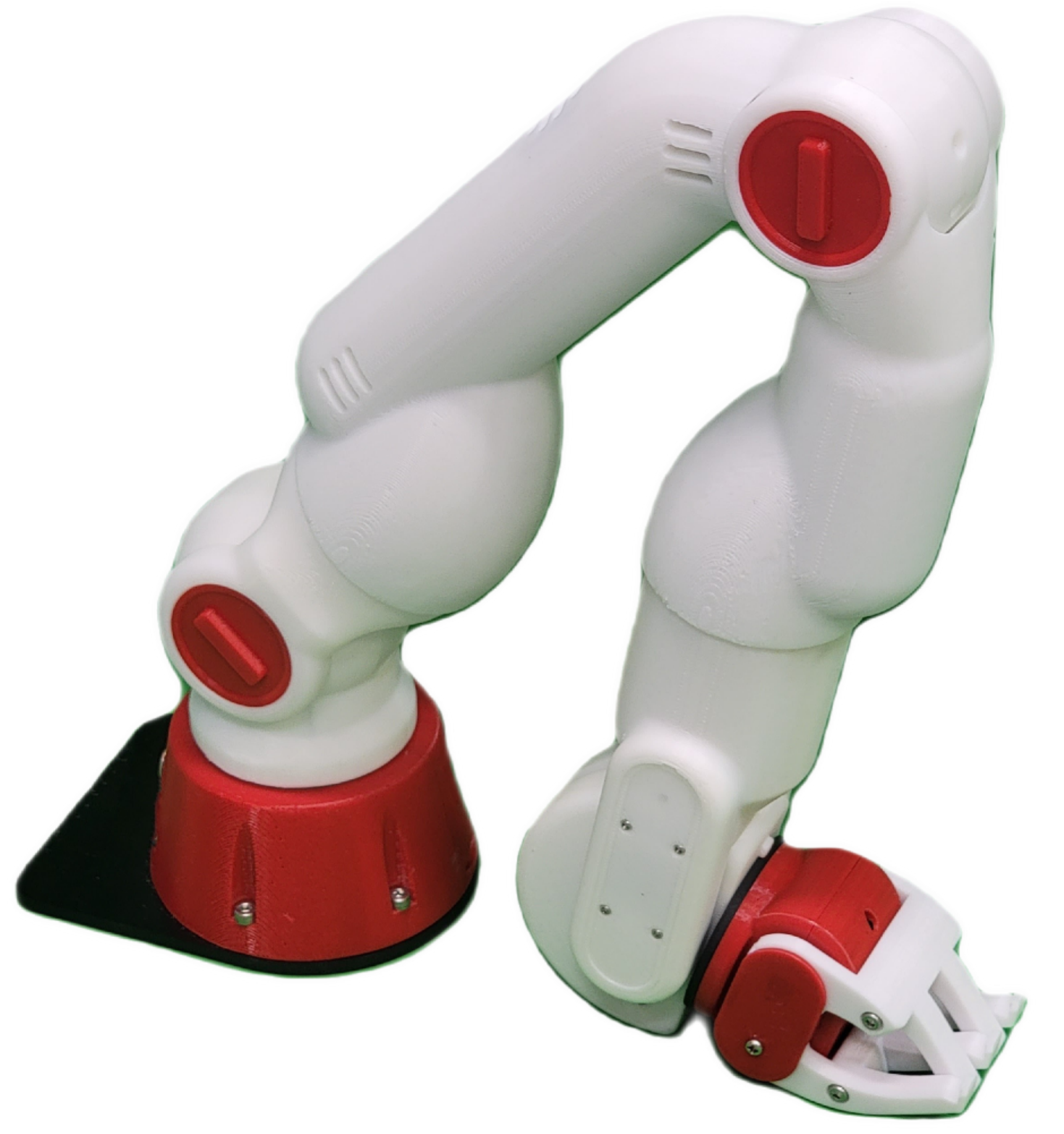}
        \caption{CRANE-X7 (RT)}
        \label{fig:CRANE-X7}
\end{figure}

Fig.~\ref{fig:CRANE-X7} illustrates a manipulator, called CRANE-X7, used in the experiments. CRANE-X7 consists of an arm with seven-DoF and a gripper with one-DoF. This robot has torque control and angle sensors on all joints and the gripper. We adopted the cross-type hand~\cite{Yamane_grasp_hand} for the gripper. In the following, joint numbers are assigned in order from the base of the robot. The gripper is marked as the eighth joint. The controller and hardware parameters are shown in Table~\ref{tab:parameters}

We assumed the dynamics of the manipulator as follows:

\begin{equation}
    \bm{J} \ddot{\bm{\theta}}^{res} = \bm{\tau}^{ref} - \bm{\tau}^{res} - \bm{D} \dot{\bm{\theta}}^{res} - \bm{g}
\end{equation}
where $\bm{J}$ is the inertia matrix, $\ddot{\bm{\theta}}_{res}$ is the response vector of the angular acceleration, $\bm{\tau}^{ref}$ is the control input vector,  $\bm{\tau}^{res}$ is the reaction force vector, $\bm{D}$ is the viscous friction coefficient matrix, and $\bm{g}$ is the gravity vector. 

The inertia matrix $\bm{J}$ and the viscous friction coefficient matrix $\bm{D}$ were diagonal matrices and are expressed as follows:
\begin{equation}
    \bm{J} = \rm{diag}
    \begin{bmatrix}
        J_{1}, J_{2}, \cdots , J_{8}
    \end{bmatrix},
\end{equation}
\begin{equation}
    \bm{D} = \rm{diag}
    \begin{bmatrix}
        D_{1}, D_{2}, \cdots , D_{8}
    \end{bmatrix}.
\end{equation}
The inertias of the first three joints $J_{1}$, $J_{2}$, and $J_{3}$ are expressed as follows:
\begin{equation}
\begin{split}
J_{1} &= 
c_{2}^{2}m_{2}\sin^{2}{\theta_{2}} + c_{3}^{2}m_{3}\sin^{2}{\theta_{2}} + 2c_{3}l_{2}m_{3}\sin^{2}{\theta_{2}} \\ 
&- \frac{1}{8}c_{4}^{2}m_{4}\{\cos(-2\theta_{2} + \theta_{3} + 2\theta_{4}) - \cos(2\theta_{2} - \theta_{3} + 2\theta_{4}) \\
&+ \cos(2\theta_{2} + \theta_{3} - 2\theta_{4}) - \cos(2\theta_{2} + \theta_{3} + 2\theta_{4})\} \\ 
&+ c_{4}^{2}m_{4}\sin^{2}{\theta_{2}}\sin^{2}{\theta_{3}}\sin^{2}{\theta_{4}} - 2c_{4}^{2}m_{4}\sin^{2}{\theta_{2}}\sin^{2}{\theta_{4}} \\ 
&+ c_{4}^{2}m_{4}\sin^{2}{\theta_{2}} + c_{4}^{2}m_{4}\sin^{2}{\theta_{4}} - 2c_{4}l_{2}m_{4}\sin^{2}{\theta_{2}}\cos{\theta_{4}} \\
&+ 2c_{4}l_{2}m_{4}\sin{\theta_{2}}\sin{\theta_{4}}\cos{\theta_{2}}\cos{\theta_{3}} + I_{x,2}\sin^{2}{\theta_{2}}\\ 
&+ I_{x,3}\sin^{2}{\theta_{2}} - I_{y,2}\sin^{2}{\theta_{2}} + I_{y,2} + l_{2}^{2}m_{3}\sin^{2}{\theta_{2}} \\
&+ l_{2}^{2}m_{4}\sin^{2}{\theta_{2}}, \\
J_{2} &= 
c_{2}^{2}m_{2} + c_{3}^{2}m_{3} + 2c_{3}l_{2}m_{3} - c_{4}^{2}m_{4}\sin^{2}{\theta_{3}}\sin^{2}{\theta_{4}} \\
&+ c_{4}^{2}m_{4} - 2c_{4}l_{2}m_{4}\cos{\theta_{4}} + I_{x,2} + I_{x,3} + l_{2}^{2}m_{3} + l_{2}^{2}m_{4}, \\
J_{3} &= 
c_{4}^{2}m_{4}\sin^{2}{\theta_{4}}
\end{split}
\end{equation}
where $m_{\bigcirc}$ is the weight of the link, $c_{\bigcirc}$ is the distance from the joint to the center of mass of the link, $l_{2,4}$ is the distance the second joint to the fourth joint, and $I_{\bigcirc,\bigcirc}$ is the inertia of the link.

The response vector of the angular acceleration $\ddot{\bm{\theta}}^{res}$ is $\begin{bmatrix}\ddot{\theta}_{1}^{res}, \ddot{\theta}_{2}^{res}, \cdots ,\ddot{\theta}_{8}^{res}\end{bmatrix}^{T}$ where the second subscript indicates the joint number. Much the same is true of the angle, the angler velocity, and reaction torque.

\begin{table}[t]
    \centering
    \caption{Controller and Hardware Parameters}
    \label{tab:parameters}
    \begin{tabular}{cccc}
    \hline
    & Parameter & Value & Unit \\ \hline\hline

    $K_{p}$ & Position P gain of all joints & 256 \\
    $K_{d}$ & Position D gain of all joints & 32.0 \\
    $K_{f}$ & Force P gain of all joints & 0.7 \\
    
    $m_{2}$ & Mass of 2nd link & 0.0128 & kg \\
    $m_{3}$ & Mass of 3rd link & 0.0000 & kg \\
    $m_{4}$ & Mass of 4th link & 0.4505 & kg \\
    $c_{2}$ & Center of mass of 2nd link & 0.0002 & m \\
    $c_{3}$ & Center of mass of 3rd link & 0.0262 & m \\
    $c_{4}$ & Center of mass of 4th link & 0.2865 & m \\
    $l_{2,4}$ & Length between the 2nd and 4th joints & 0.2500 & m \\

    $I_{x,2}$ & Inertia of 2nd link around x-axis & 0.0197 & kg~m$^2$ \\
    $I_{x,3}$ & Inertia of 3rd link around x-axis & 0.0197 & kg~m$^2$ \\
    $I_{y,2}$ & Inertia of 2nd link around y-axis & 0.0008 & kg~m$^2$ \\

    $J_{4}$ & Inertia of 4th joint & 0.0370 & kg~m$^2$ \\
    $J_{5}$ & Inertia of 5th joint & 0.0054 & kg~m$^2$ \\
    $J_{6}$ & Inertia of 6th joint & 0.0066 & kg~m$^2$ \\
    $J_{7}$ & Inertia of 7th joint & 0.0049 & kg~m$^2$ \\
    $J_{8}$ & Inertia of 8th joint & 0.0055 & kg~m$^2$ \\

    $D_{1}$ & Viscous friction coefficient of 1st joint & 0.0443 & N~m~s\slash rad \\
    $D_{2}$ & Viscous friction coefficient of 2nd joint & 0.2343 & N~m~s\slash rad \\
    $D_{3}$ & Viscous friction coefficient of 3rd joint & 0.0501 & N~m~s\slash rad \\
    $D_{4}$ & Viscous friction coefficient of 4th joint & 0.1820 & N~m~s\slash rad \\
    $D_{5}$ & Viscous friction coefficient of 5th joint & 0.0122 & N~m~s\slash rad \\
    $D_{6}$ & Viscous friction coefficient of 6th joint & 0.0196 & N~m~s\slash rad \\
    $D_{7}$ & Viscous friction coefficient of 7th joint & 0.0170 & N~m~s\slash rad \\
    $D_{8}$ & Viscous friction coefficient of 8th joint & 0.0105 & N~m~s\slash rad \\

    $f_{C1}$ & Cutoff frequency of 1st joint & 10.0 & rad\slash s \\
    $f_{C2}$ & Cutoff frequency of 2nd joint & 15.0 & rad\slash s \\
    $f_{C3}$ & Cutoff frequency of 3rd joint & 10.0 & rad\slash s \\
    $f_{C4}$ & Cutoff frequency of 4th joint & 15.0 & rad\slash s \\
    $f_{C5}$ & Cutoff frequency of 5th joint & 90.0 & rad\slash s \\
    $f_{C6}$ & Cutoff frequency of 6th joint & 90.0 & rad\slash s \\
    $f_{C7}$ & Cutoff frequency of 7th joint & 90.0 & rad\slash s \\
    $f_{C8}$ & Cutoff frequency of 8th joint & 90.0 & rad\slash s \\
    
    \hline
    \end{tabular}
\end{table}

The gravity vector is expressed as follows:
\begin{equation}
    \bm{g} = 
    \begin{bmatrix}
        0, g_{2}, g_{3}, g_{4}, 0, 0, 0, 0,
    \end{bmatrix}^{T}
\end{equation}
where
\begin{equation}
\begin{split}
g_{2} &= 
g[c_{2}m_{2}\sin{\theta_{2}} + m_{3}(c_{3} + l_{2})\sin{\theta_{2}} \\
&+ m_{4}\{c_{4}\sin{\theta_{4}}\cos{\theta_{2}}\cos{\theta_{3}} \\
&+ (-c_{4}\cos{\theta_{4}} + l_{2} + l_{3})\sin{\theta_{2}}\}], \\
g_{3} &= 
-c_{4}gm_{4}\sin{\theta_{2}}\sin{\theta_{3}}\sin{\theta_{4}}, \\
g_{4} &=
c_{4}gm_{4}(\sin{\theta_{2}}\cos{\theta_{3}}\cos{\theta_{4}} - \sin{\theta_{4}}\cos{\theta_{2}}).
\end{split}    
\end{equation}

\begin{figure}[t]
    \centering
    \includegraphics{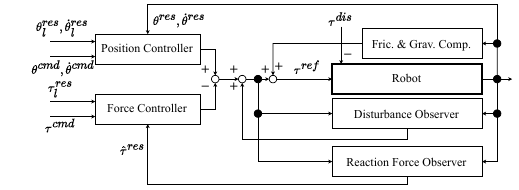}
        \caption{Block Diagram of Controller}
        \label{Fig:Block Diagram of Controller}
\end{figure}

\begin{figure*}[t]
\centering
\begin{minipage}[c]{0.3\linewidth}
    \centering
    \includegraphics[width=0.7\columnwidth]{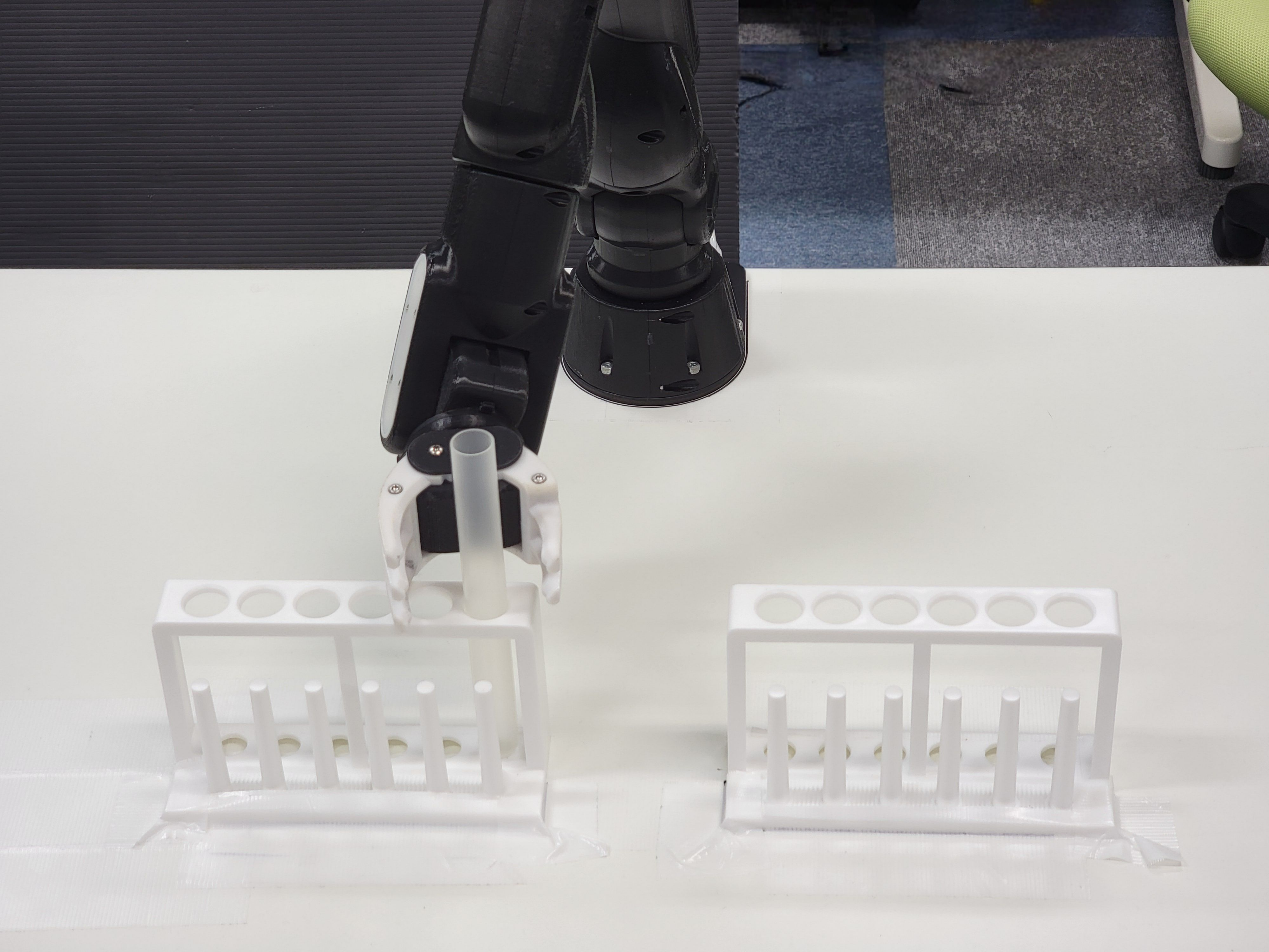}
    \subcaption{Picking Test Tube}
\end{minipage}
    \begin{minipage}[c]{0.03\linewidth}
        \centering
        \includegraphics[width=0.9\columnwidth]{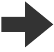}
    \end{minipage}
\begin{minipage}[c]{0.3\linewidth}
    \centering
    \includegraphics[width=0.7\columnwidth]{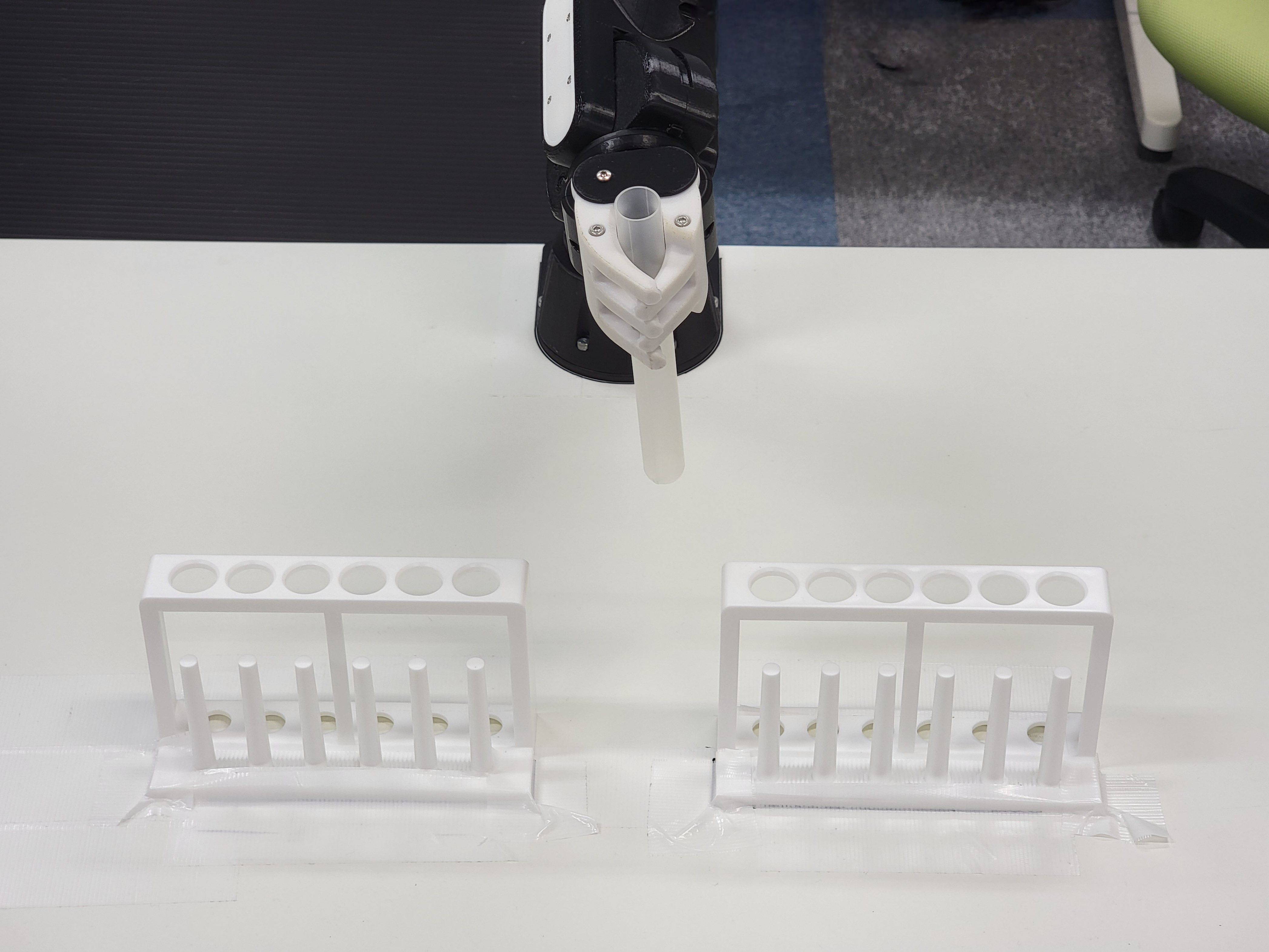}
    \subcaption{Moving Test Tube}
\end{minipage}
    \begin{minipage}[c]{0.03\linewidth}
        \centering
        \includegraphics[width=0.9\columnwidth]{Fig/arrow.pdf}
    \end{minipage}
\begin{minipage}[c]{0.3\linewidth}
    \centering
    \includegraphics[width=0.7\columnwidth]{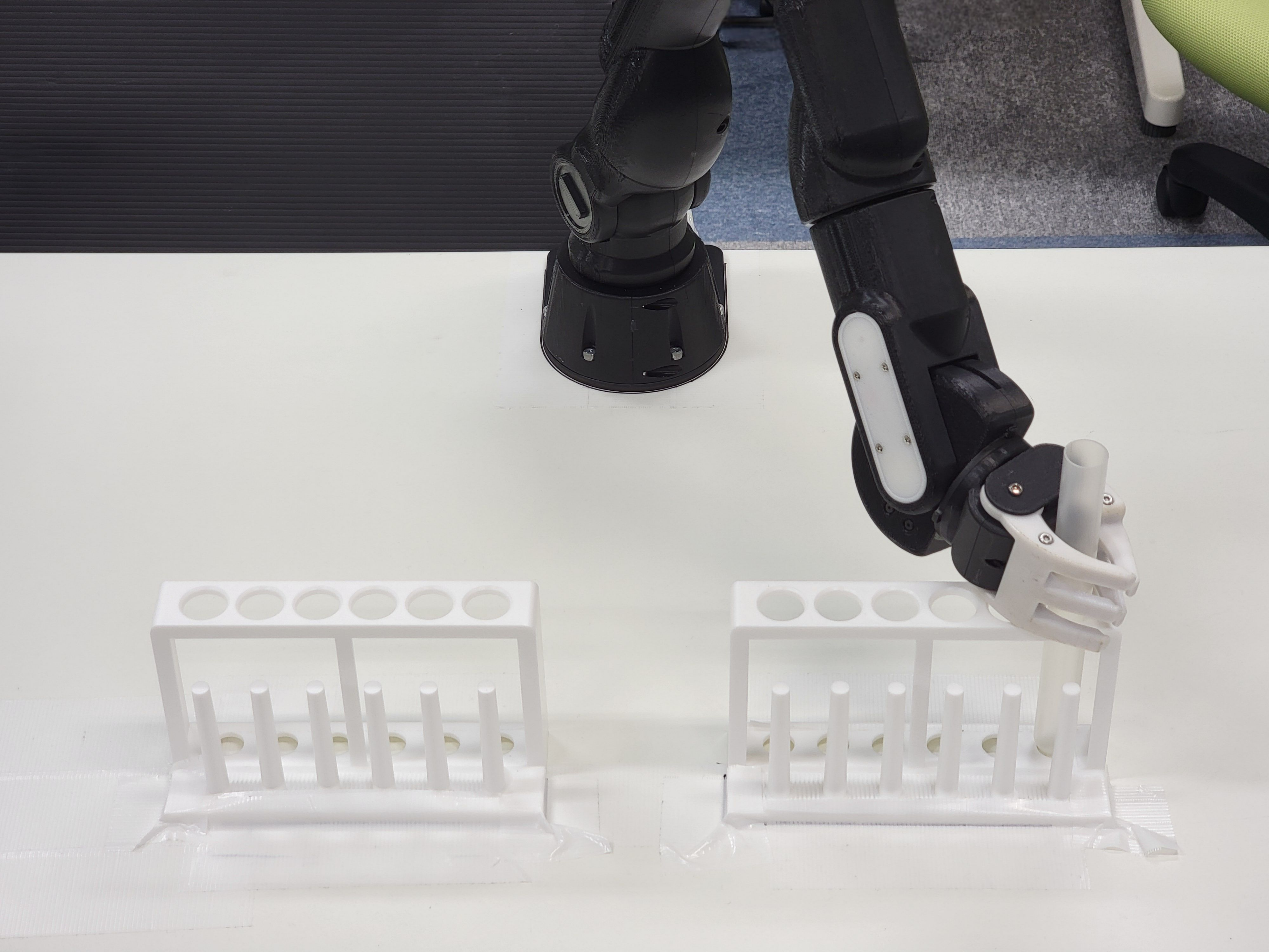}
    \subcaption{Inserting Test Tube}
\end{minipage}
\caption{Snapshots of Motion}
\label{Fig:Snapshot of Motion}
\end{figure*}

The block diagram of the controller is shown in Fig.~\ref{Fig:Block Diagram of Controller}. The disturbance is represented by $\bm{\tau}^{dis}$. 
We used a disturbance observer (DOB)~\cite{Ohnishi_DOB}\cite{Sariyildiz_DOB35} and a reaction force observer (RFOB)~\cite{Murakami_RFOB}. We explicitly compensated friction and gravity. In addition, we suppressed other unknown disturbances by using the DOB. The sum of friction compensation, gravity compensation, and output of DOB is represented by $\hat{\bm{\tau}}^{dis}$. The reaction force was estimated by the RFOB. The estimated reaction force is represented by $\hat{\bm{\tau}}^{res}$. Moreover, angular velocities were calculated by pseudo-differentiation using the values obtained from the angle sensor. The command values, $\bm{\theta}^{cmd}$, $\dot{\bm{\theta}}^{cmd}$, and $\bm{\tau}^{cmd}$, which were input to the position controller and force controller, were the editor response values for the follower robot controller or the follower response value for the editor robot controller.

In the position controller, we used a proportional-derivative (PD) controller of position. In the force controller, we used a proportional (P) controller of force. The gain of the PD and the P controller, $K_{p}$, $K_{d}$, and $K_{f}$, were determined manually.

To estimate the inertia, the viscous friction, and the gravity, we used unilateral control to collect data and calculate inverse dynamics, based on the method~\cite{Inami_SystemIdentification}. This method allowed system identification at frequencies and operating ranges closer to practical use. Finally, we performed manual adjustments. 

\begin{figure*}[t]
\centering
\begin{minipage}[b]{0.49\linewidth}
    \centering
    \includegraphics[width=0.65\columnwidth]{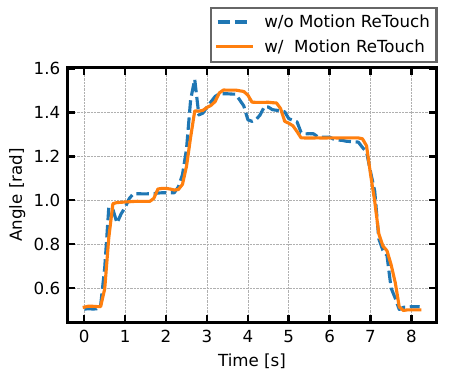}
    \caption{Angle of 4th Joint}
    \label{Fig:Angle}
\end{minipage}
\begin{minipage}[b]{0.49\linewidth}
    \centering
    \includegraphics[width=0.65\columnwidth]{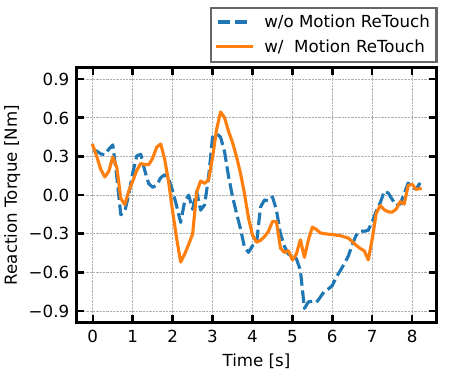}
    \caption{Reaction Torque of 4th Joint}
    \label{Fig:ReactionForce}
\end{minipage}
\end{figure*}

In Motion ReTouch, the control inputs $\bm{\tau}^{ref}$ of the follower and editor robots were given by the following equations:
\begin{equation}
\begin{split}
    \bm{\tau}_{f}^{ref} =&
         \frac{\bm{J}}{2} (K_{p} + s K_{d})\{ \alpha\bm{\theta}_{e}^{res} + (1-\alpha)\bm{\theta}_{l}^{res}-\bm{\theta}_{f}^{res})\} \\
         -& \frac{1}{3} K_f (\bm{\tau}_{l}^{res} + \bm{\tau}_{e}^{res} + \bm{\tau}_{f}^{res}) + \hat{\bm{\tau}}_{f}^{dis},
\end{split}
\label{Eq:tau_f^ref}
\end{equation}
\begin{equation}
\begin{split}
    \bm{\tau}_{e}^{ref} =&
         \frac{\bm{J}}{2} (K_{p} + s K_{d})\{ \alpha\bm{\theta}_{f}^{res} + (1-\alpha)\bm{\theta}_{l}^{res}-\bm{\theta}_{e}^{res})\} \\
         -& \frac{1}{3} K_f (\bm{\tau}_{l}^{res} + \bm{\tau}_{f}^{res} + \bm{\tau}_{e}^{res}) + \hat{\bm{\tau}}_{e}^{dis}
\end{split}
\label{Eq:tau_e^ref}
\end{equation}
where s is the Laplace operator.

We set the parameter $\alpha$ to $0.5$. The control cycle was set to 500Hz, which was the fastest control cycle of this manipulator. The cutoff frequencies of the low-pass filters of the DOB, the RFOB, and the pseudo-differential were the same, and their values are $f_{C1}, f_{C2}, \cdots, f_{C8}$. We used the same parameters in four-channel bilateral control, Motion ReTouch, and motion-copying.


\subsection{TASK}

In this paper, we evaluated the proposed method by moving test tubes from one test tube rack to another. The snapshot of the motion is shown in Fig.~\ref{Fig:Snapshot of Motion}. This task involved some contact when removing and inserting test tubes. The test tube used was made of polypropylene and had a diameter of 18 mm and a length of 180 mm. The test tube rack used had a hole diameter of 19 mm and a height of 94 mm. The test tube racks were fixed to the desk.

We defined failure as missing or dropping the test tube, failing to insert the test tube, or inserting the test tube at an angle.

\subsection{EXPERIMENTAL STEPS}
First, we collected motion data through four-channel bilateral control. In this step, the robot which humans operated was called a leader robot, the other robot which followed the leader robot was called a follower robot. At this time, the command values of the follower robot were recorded.

To make the movements more difficult, the motion data was sped up by three times. This was achieved by skipping every two lines of the time series data and tripling only the angular velocity command values of the follower robot.

Second, we executed Motion ReTouch with the leader robot (motion data), the follower robot, and the editor robot. The command values of the follower robot were recorded.

Finally, we evaluated the task success rate and the robot's behavior with and without Motion ReTouch. We performed motion-copying 10 times with and without Motion ReTouch. We also recorded the response value of the follower robot at that time.


\section{RESULT}

It took 24 seconds to transfer the tubes using four-channel bilateral control. We sped up the data by 3x, hence the data was about 8 seconds.

The success rate of the experiment is shown in Table~\ref{tab:success rate}. The success rate of original speed data was 100\%. However, because we sped up the data by 3x, the success rate fell to 0\%. By using Motion ReTouch to correct the 3x speed data, we were able to increase the success rate to 100\%. It is noted that all of the mistakes were caused by putting the test tubes in the wrong position.


\begin{table}[t]
    \centering
    \caption{Success Rate of Traspoting Test Tube}
    \label{tab:success rate}
    \begin{tabular}{cccc}
      & 
        \begin{tabular}{c}
        Original Speed \\ Motion
        \end{tabular} & 
        \begin{tabular}{c}
        3x Speed \\
        w\slash o Motion \\ ReTouch
        \end{tabular} &
        \begin{tabular}{c}
        3x Speed \\
        w\slash \ Motion \\ ReTouch
        \end{tabular}
         \\ \hline
    Success Rate 
     & 10\slash10  & 0\slash10 & 10\slash10
    \end{tabular}
\end{table}

The response values of position and reaction torque of the fourth joint in motion copying are shown in Fig.~\ref{Fig:Angle} and Fig.~\ref{Fig:ReactionForce}, respectively. The mean values obtained from 10 samples are shown. The motion data which was sped up by 3x is represented by the dashed blue line and The motion data which was sped up by 3x and was modified with Motion ReTouch is represented by the orange line. The variance of the 10 samples for both position and force was small.

The position response values did not differ significantly between the two models. This is because they were based on the same motion data, and the position information was not significantly rewritten in Motion ReTouch. Although the position trajectory had become slightly smoother with Motion ReTouch. This may be due to the addition of the editor robot and the human operating it, which may have difficulty tracking high-frequency movements.

On the other hand, the reaction torque values were drastically changed in some places with Motion ReTouch. In particular, we can see that the reaction force of the model with Motion ReTouch is stable between 5 and 7 seconds. This was the time when the robot was searching for a place to insert the test tube. According to the reaction torque values, stabilizing the reaction force when inserting the test may contribute to the improved success rate. On the other hand, there were times when the force information did not change. This is because the force data was not changed from the original data when the editor robot was not touched.


\section{CONCLUSION AND LIMITATION}

In this paper, we proposed Motion ReTouch to modify the data earned with four-channel bilateral control. Motion ReTouch was a combination of multilateral control and a motion-copying system. We realized Motion ReTouch by combining three robots: a virtual leader robot which reproduces motion data obtained by past bilateral control, a follower robot which performs tasks, and an editor robot which makes changes. Motion ReTouch made it possible to change not only position information but also force information. We could improve the success rate in the test tube transfer task by Motion ReTouch.

However, Motion ReTouch has several limitations. First, Motion ReTouch is unable to make any major changes to the position of the follower robot. This is because one differential mode was independent of the others, and there was always a force pulling it back to the original trajectory, and the force increased as the change became larger. As proposed in~\cite{takahashi_CHG-DAgger}, changing the leader robot to a learned policy can avoid this problem. Second, the operators need to remember the original data of the position and force to modify them. The position is relatively easy to remember, but the force information is difficult to remember because it is not visible. Finally, we are not yet able to make corrections in the time domain, and if the corrections are not made at the right time, there is a possibility that the data will be worse. In our future research, improvements are needed to allow for correction in the time direction. Modifications will be easier when only complex movements are performed slowly.

\bibliography{references}

\begin{thebibliography}{10}
\providecommand{\url}[1]{#1}
\csname url@samestyle\endcsname
\providecommand{\newblock}{\relax}
\providecommand{\bibinfo}[2]{#2}
\providecommand{\BIBentrySTDinterwordspacing}{\spaceskip=0pt\relax}
\providecommand{\BIBentryALTinterwordstretchfactor}{4}
\providecommand{\BIBentryALTinterwordspacing}{\spaceskip=\fontdimen2\font plus
\BIBentryALTinterwordstretchfactor\fontdimen3\font minus \fontdimen4\font\relax}
\providecommand{\BIBforeignlanguage}[2]{{%
\expandafter\ifx\csname l@#1\endcsname\relax
\typeout{** WARNING: IEEEtran.bst: No hyphenation pattern has been}%
\typeout{** loaded for the language `#1'. Using the pattern for}%
\typeout{** the default language instead.}%
\else
\language=\csname l@#1\endcsname
\fi
#2}}
\providecommand{\BIBdecl}{\relax}
\BIBdecl

\bibitem{Hussein_ImitationLearningSurvey}
A.~Hussein, M.~M. Gaber, E.~Elyan, and C.~Jayne, ``Imitation learning: A survey of learning methods,'' \emph{ACM Computing Surveys (CSUR)}, vol.~50, no.~2, pp. 1--35, 2017.

\bibitem{fang_SurveyImitationLearning}
B.~Fang, S.~Jia, D.~Guo, M.~Xu, S.~Wen, and F.~Sun, ``Survey of imitation learning for robotic manipulation,'' \emph{International Journal of Intelligent Robotics and Applications}, vol.~3, pp. 362--369, 2019.

\bibitem{Attia_OverviewImitationLearinig}
A.~Attia and S.~Dayan, ``Global overview of imitation learning,'' \emph{arXiv preprint arXiv:1801.06503}, 2018.

\bibitem{gavspar_DynamicMovementPrimitives}
T.~Ga{\v{s}}par, B.~Nemec, J.~Morimoto, and A.~Ude, ``Skill learning and action recognition by arc-length dynamic movement primitives,'' \emph{Robotics and autonomous systems}, vol. 100, pp. 225--235, 2018.

\bibitem{zhao_ALOHA}
T.~Z. Zhao, V.~Kumar, S.~Levine, and C.~Finn, ``Learning fine-grained bimanual manipulation with low-cost hardware,'' \emph{arXiv preprint arXiv:2304.13705}, 2023.

\bibitem{Fu_MobileALOHA}
Z.~Fu, T.~Z. Zhao, and C.~Finn, ``Mobile aloha: Learning bimanual mobile manipulation with low-cost whole-body teleoperation,'' \emph{arXiv preprint arXiv:2401.02117}, 2024.

\bibitem{Adachi_imitate_bilate}
T.~Adachi, K.~Fujimoto, S.~Sakaino, and T.~Tsuji, ``Imitation learning for object manipulation based on position/force information using bilateral control,'' in \emph{2018 IEEE/RSJ International Conference on Intelligent Robots and Systems (IROS)}, 2018, pp. 3648--3653.

\bibitem{Buamanee_BiACT}
T.~Buamanee, M.~Kobayashi, Y.~Uranishi, and H.~Takemura, ``Bi-act: Bilateral control-based imitation learning via action chunking with transformer,'' in \emph{2024 IEEE International Conference on Advanced Intelligent Mechatronics (AIM)}, 2024, pp. 410--415.

\bibitem{Sakaino_bilate_obliquecordinate}
S.~Sakaino, T.~Sato, and K.~Ohnishi, ``Multi-dof micro-macro bilateral controller using oblique coordinate control,'' \emph{IEEE Transactions on Industrial Informatics}, vol.~7, no.~3, pp. 446--454, 2011.

\bibitem{sasagawa_cooperation}
A.~Sasagawa, K.~Fujimoto, S.~Sakaino, and T.~Tsuji, ``Imitation learning based on bilateral control for human--robot cooperation,'' \emph{IEEE Robotics and Automation Letters}, vol.~5, no.~4, pp. 6169--6176, 2020.

\bibitem{Yamane_grasp_hand}
K.~Yamane, Y.~Saigusa, S.~Sakaino, and T.~Tsuji, ``Soft and rigid object grasping with cross-structure hand using bilateral control-based imitation learning,'' \emph{IEEE Robotics and Automation Letters}, vol.~9, no.~2, pp. 1198--1205, 2024.

\bibitem{Akagawa_BiIL}
T.~Akagawa and S.~Sakaino, ``Autoregressive model considering low frequency errors in command for bilateral control-based imitation learning,'' \emph{IEEJ Journal of Industry Applications}, vol.~12, no.~1, pp. 26--32, 2023.

\bibitem{Katsura_Multilateral}
S.~Katsura, Y.~Matsumoto, and K.~Ohnishi, ``Realization of "law of action and reaction" by multilateral control,'' \emph{IEEE Transactions on Industrial Electronics}, vol.~52, no.~5, pp. 1196--1205, 2005.

\bibitem{Yokokura_MotionCopyingStability}
Y.~Yokokura, S.~Katsura, and K.~Ohishi, ``Stability analysis and experimental validation of a motion-copying system,'' \emph{IEEE Transactions on Industrial Electronics}, vol.~56, no.~10, pp. 3906--3913, 2009.

\bibitem{igarashi_motioncopy}
K.~Igarashi and S.~Katsura, ``Motion-data processing and reproduction based on motion-copying system,'' \emph{IEEJ Journal of Industry Applications}, vol.~4, no.~5, pp. 543--549, 2015.

\bibitem{Fujisaki_MotionCopying}
K.~Fujisaki and S.~Katsura, ``Motion-copying system with in-tool sensing,'' \emph{IEEJ Journal of Industry Applications}, vol.~12, no.~4, pp. 793--799, 2023.

\bibitem{Zhang_ImitationVirtualReality}
T.~Zhang, Z.~McCarthy, O.~Jow, D.~Lee, X.~Chen, K.~Goldberg, and P.~Abbeel, ``Deep imitation learning for complex manipulation tasks from virtual reality teleoperation,'' in \emph{2018 IEEE international conference on robotics and automation (ICRA)}.\hskip 1em plus 0.5em minus 0.4em\relax IEEE, 2018, pp. 5628--5635.

\bibitem{nishimura_HapticSharedControl}
R.~Nishimura, T.~Wada, and S.~Sugiyama, ``Haptic shared control in steering operation based on cooperative status between a driver and a driver assistance system,'' \emph{J. Hum.-Robot Interact.}, vol.~4, no.~3, p. 19–37, Dec. 2015.

\bibitem{dragan_PolicyBlending}
A.~D. Dragan and S.~S. Srinivasa, ``A policy-blending formalism for shared control,'' \emph{The International Journal of Robotics Research}, vol.~32, no.~7, pp. 790--805, 2013.

\bibitem{muelling_SharedAutonomy}
K.~Muelling, A.~Venkatraman, J.-S. Valois, J.~Downey, J.~Weiss, S.~Javdani, M.~Hebert, A.~B. Schwartz, J.~L. Collinger, and J.~A. Bagnell, ``Autonomy infused teleoperation with application to bci manipulation,'' \emph{arXiv preprint arXiv:1503.05451}, 2015.

\bibitem{sun_BilateSharedControl}
D.~Sun and Q.~Liao, ``Asymmetric bilateral telerobotic system with shared autonomy control,'' \emph{IEEE Transactions on Control Systems Technology}, vol.~29, no.~5, pp. 1863--1876, 2020.

\bibitem{ross_dagger}
S.~Ross and D.~Bagnell, ``Efficient reductions for imitation learning,'' in \emph{Proceedings of the thirteenth international conference on artificial intelligence and statistics}.\hskip 1em plus 0.5em minus 0.4em\relax JMLR Workshop and Conference Proceedings, 2010, pp. 661--668.

\bibitem{kelly_hg-dagger}
M.~Kelly, C.~Sidrane, K.~Driggs-Campbell, and M.~J. Kochenderfer, ``Hg-dagger: Interactive imitation learning with human experts,'' in \emph{2019 International Conference on Robotics and Automation (ICRA)}.\hskip 1em plus 0.5em minus 0.4em\relax IEEE, 2019, pp. 8077--8083.

\bibitem{takahashi_CHG-DAgger}
T.~Takahashi, ``Chg-dagger: Interactive imitation learning with human-policy cooperative control,'' in \emph{CoRL 2024 Workshop CoRoboLearn: Advancing Learning for Human-Centered Collaborative Robots}, 2024.

\bibitem{MultiTeletrain_Katsura2008}
S.~Katsura, T.~Suzuyama, and K.~Ohishi, ``Force transmission control in multilateral system for teletraining,'' \emph{IEEJ Transactions on Industry Applications}, vol. 128, no.~6, pp. 826--832, 2008.

\bibitem{Ohnishi_DOB}
K.~Ohnishi, M.~Shibata, and T.~Murakami, ``Motion control for advanced mechatronics,'' \emph{IEEE/ASME Transactions on Mechatronics}, vol.~1, no.~1, pp. 56--67, 1996.

\bibitem{Sariyildiz_DOB35}
E.~Sariyildiz, R.~Oboe, and K.~Ohnishi, ``Disturbance observer-based robust control and its applications: 35th anniversary overview,'' \emph{IEEE Transactions on Industrial Electronics}, vol.~67, no.~3, pp. 2042--2053, 2020.

\bibitem{Murakami_RFOB}
T.~Murakami, F.~Yu, and K.~Ohnishi, ``Torque sensorless control in multidegree-of-freedom manipulator,'' \emph{IEEE Transactions on Industrial Electronics}, vol.~40, no.~2, pp. 259--265, 1993.

\bibitem{Inami_SystemIdentification}
K.~Inami, K.~Yamane, and S.~Sakaino, ``Loss function considering dead zone for neural networks,'' in \emph{2024 IEEE 18th International Conference on Advanced Motion Control (AMC)}, 2024, pp. 1--6.

\end{thebibliography}
\bibliographystyle{IEEEtran}

\end{document}